# Vision Based Railway Track Monitoring Using Deep Learning

Shruti Mittal
*GE Transportation*
Bangalore, India
shruti.mitttal@ge.com

Dattaraj Rao
*GE Transportation*
Bangalore, India
dattaraj.rao@ge.com

**Computer vision based methods have been explored in the past for detection of railway track defects, but full automated surveillance has always been a challenge because both traditional image processing methods and deep learning classifiers trained from scratch fail to generalize that well to infinite novel scenarios seen in the real world, given limited amount of labeled data. Advancements have been made recently to make machine learning models utilize knowledge from a different but related domain. In this paper, we show that even though similar domain data is not available, transfer learning provides the model understanding of other real-world objects and enables training production scale deep learning classifiers for uncontrolled real-world data. Our models efficiently detect both track defects like sunkinks, loose ballast and railway assets like switches and signals. Models were validated with hours of track videos recorded in different continents resulting in different weather conditions, different ambience and surroundings. Different defects detected contribute to a track health index which can be used to monitor complete rail network.**

## I. INTRODUCTION

Railway network is one of the major components of any country's infrastructure. Improper maintenance of the track infrastructure can cause major revenue loss and have bigger implications like loss of life due to accidents. At present, all the maintenance work is done manually by sending a dedicated team for rail inspection every few days. Railroad companies also need to maintain and keep updating GPS database of assets like signals, switches and mileposts to implement temporary speed restrictions between specific start and end points, enable railroad interoperability. Using high definition videos recorded by train passing through or by drones eliminates the need for repetitive manual inspection and alerts can be provided to maintenance crew to plan for track and asset maintenance in the region with high probability of damage. Onboard detection of these railway assets can also be used in operator assistance systems and onboard track defect detection or obstruction detection can help in early warning systems.

Traditional image processing methods fail to accommodate variability in track seen and gives a lot of false positives. For example, a system trained to identify sunkinks or bent rails get affected by switches, double rails, support rails, and any straight line near track like a water pipe going along with track or platform edge at a station. Deep learning classifiers trained from scratch need a huge amount of data. Other than being expensive to label all the data, it is also difficult to obtain that many images of defective track because it is a rare scenario, as it should be! As a result, this works well on the cleaned up, manually filtered and constructed dataset. They even sometimes work on railway tracks present in deserted areas where there are limited distractions which could be covered in the negative dataset but they tend to fail whenever it sees something new. If there are tracks going through the city or through a region with a different kind of scenery than that covered in training dataset, these models tend to fail.

This brings into picture transfer learning. Classifiers proposed in this paper have been pretrained on ImageNet and COCO dataset before being trained on railway data. Being trained on millions of images from these datasets, models gain knowledge about thousands of different objects seen in real life. They learn about different backgrounds and when fine-tuned on relevant data, understands its features much more effectively. They don't get confused easily with other objects seen around and hence give much less false positives.

## II. RELATED WORKS

At present, all the maintenance work is done manually by sending a dedicated team for rail inspection every few days. A complete new hardware system with a lot of sensors can automate a lot of this work but these setups require a separate team to go on the tracks to be inspected with the required instruments. A few vision based methods like by Karakose et al, 2017 [5]; Vijaykumar et al, 2015 [8] have also been explored for rail track defect detection. But due to lack of sufficient labeled data they have stayed limited to conventional image processing methods. The defects like sunkinks or broken track are seen rarely, however, for training a convolutional classifier, it needs thousands of images of each category. Also, labeling all the video data for this is a very time-consuming task. Both these issues have resulted in limited exploratory studies like in Faghih-Roohi et al, 2016 [6], Gibert et al, 2017 [7]. Recently, techniques like data augmentation and transfer learning have shown that efficient deep learning models can be trained even with small amount of data. New models like inception have been developed by Szegedy et al, 2015 [4] to maximize use of information available from limited dataset. Also, tensorflow and keras have released models pretrained on datasets like ImageNet and COCO containing millions of labelled images enabling training of new models using limited computational resources.

## III. BACKGROUND

In recent times, convolutional neural networks have become quite popular for image related tasks. Unlike conventional neural networks, their 2D architecture helps them to extract localized features of images well and their weight sharing provides translational invariance for structures detected along with protecting them from overfitting and enabling training with limited computing resources. Stacking those convolutional layers, enables the model to learn more complicated features. CNNs have proved that given sufficient data, the task of image classification with accuracy similar to or even at times better than that of humans is no longer hard for computers. These CNNs have been improved upon in a lot of ways to use fewer parameters to be more computationally efficient and avoid overfitting, to learn much more complicated features by going deeper and to handle tasks of multiple object detection and segmentation in addition to image classification by using region proposal networks.

### A. Inception

Google came up with the Inception module Szegedy et al in 2015 [4] in which they proposed, instead of stacking convolutional layers in a sequential manner, if convolutions with different filter sizes can be processed in parallel and then concatenated and fed to next layer, model will be retaining a lot of useful information about features of different sizes, which would be lost otherwise. The computational burden added due to this was compensated for by using 1x1 convolutions which enabled controlling number of input channels for the next layer without losing important spatial information and without increasing number of trainable parameters. These techniques resulted in models which were computationally very efficient and giving state of the art accuracy.

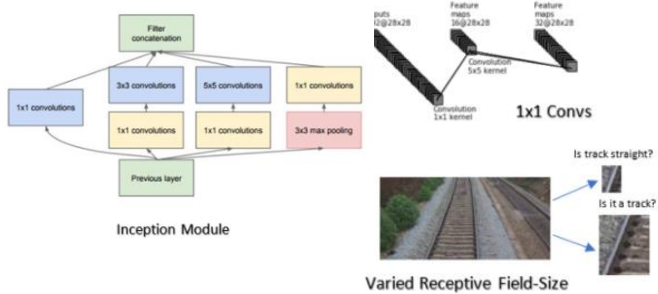

**Figure 1: Inception Model[1]\***

### B. ResNet

Residual Learning framework was introduced by Microsoft He et al, in late 2015 [2]. The basic idea behind it is that the inputs of a lower layer are added to the inputs of a higher layer due to which higher layers make inference based on both the extracted feature maps and the original inputs. It was observed that due to this change, now the layers only need to learn a residual function instead of an unreferenced function or in other words, layers are now learning only the incremental change in the original input which can help overall model to classify correctly. This also provided a highway route for the error to propagate to lower layers during back propagation enabling training of much deeper networks compared to what were possible earlier and hence a considerable gain in the model accuracy.

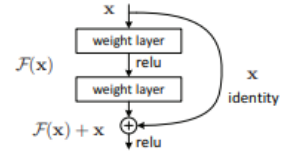

**Figure 2: Residual Net: Building Block[2]**

### C. Faster R-CNN

Due to their inherent architecture, a simple CNN model cannot tell the location of the object it detected. To take image classification to the next step of object detection and be able to draw a bounding box, region based CNNs were introduced in 2013. They use a region proposal method to identify bounding boxes with highest probability to contain an object and then classification is run for the image inside those boxes. In 2016, faster R-CNN was introduced by Ren et al, 2015 [3], which instead of using a separate region proposal method, integrated it with CNN model so that the same feature maps created can be used both for proposing bounding boxes and classification. Using CNN feature maps resulted in higher quality region proposals and greatly reduced the overall processing time.

### D. Transfer Learning

All the improvements on CNN mentioned above, have one big limitation - need of huge amount of labeled data, which is not always available in real world scenarios. This brings into picture Transfer Learning which basically refers to learning features from labeled dataset of another problem and utilizing that knowledge to solve the issue in hand. The initial layers in a CNN model extract low level features like edges, corners, pixel intensities while final layers combine this low-level information to make more problem specific feature maps so that they can execute the final task of classification or recognition. Transfer learning proposes keep the weights in lower layers intact providing the model good initial feature extractors and then fine-tune only the final layers for your specific problem with limited data.

## IV. EXPERIMENTS AND RESULTS

Currently 100 GB of full HD videos are being collected per day from cameras mounted on top of a few locomotives. These videos are used to pursue video analytics with two objectives – one is track defects detection and the other is railway assets mapping. The GPU used to run these models is Titan X. For some models, speed on CPU has also been reported. That is a system with 6 cores and 32 GB RAM.

---

[1] Image Composed using architecture module from [11]

[2] Image from [2]

## A. Track Defects

Railway track defects can be categorized as follows – one is issues with track itself like broken track or misalignment also known as sunkinks in which part of track expands due to heat and gets bent due to lack of space. Other category is issues in between the tracks like loose ballast or missing crosstie which reduces the load bearing capacity of tracks and hence that region is more likely to see a rail damage in near future. Track defects from category 1 are safety critical issues and can result in derailment. Thus, a small number of false positives are acceptable but there should not be any false negative for these. Category 2 defects provide priority to maintenance crews. So, these models should have very less false positives. However, if in a particular region with loose ballast, even if it detected some of the frames as loose ballast, it is good enough to identify the region which needs maintenance and we do not need to know one off cases. Thus, false negatives in this case are acceptable. We have trained binary classifiers for one defect in each of these categories. The classifiers were trained using cropped region of interest of the frame marked by green rectangle in the results.

### 1) Loose Ballast

A track is identified as having loose ballast when it doesn't have enough gravel distributed in between tracks making depth of cross ties visible. In case of insufficient ballast, the load of the loco doesn't get evenly distributed leading to the breakage of cross ties and eventually track.

Both resnet-50 and inception models, pretrained on ImageNet data were first tried by replacing their top layer by a two node softmax layer. The models were trained using just 181 positives and 340 negative images. These images were however augmented using mirroring, rotation, image shift and varied brightness. The trained models were tested on 25000 samples from a route covering 470 kms. Inception model claimed 88.9% of the track as having loose ballast while Resnet claimed only 1.5% of the track as defective. As explained earlier, as this is a defect that triggers maintenance in the region, even if it detects only some of the actual frames containing loose ballast, it is good enough to identify the region. Hence, false negatives are easily acceptable in this case but not a lot of false positives, making ResNet a preferred choice for this, which was optimized further.

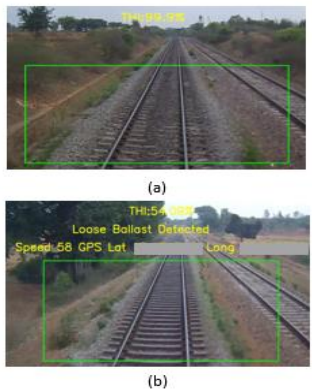

**Figure 3: (a) shows good ballast condition, (b) shows a loose ballast case being detected**

After tuning the model for meta parameters - number of trainable layers, activation function in fully connected layers, amount of regularization, the final optimized model had following architecture. The initial layers of the model were kept frozen while top ten layers were trainable. Two fully connected layers and a softmax layer were added instead of the top layer of original model. Dropout was also added before these layers. Trained model was validated on 32 hrs worth of data, spread over a month and 0.46% of the track covered was identified as having loose ballast with a precision of 96%. Recall could not be calculated for 32 hrs of data as it would need labelling of all the actual cases of loose ballast in the all the videos. As distribution of gravels is not a very straight forward feature to identify, training deeper models was helpful as it could understand more complicated features.

**Table 1: Results for Loose Ballast Model**

| Parameters | Inception-v3 Top Modified | ResNet 50 Top Modified | ResNet 50 Optimized | ResNet 50 Optimized |
|---|---|---|---|---|
| Videos Parsed | 9 hrs | 9 hrs | 9hrs | 32 hrs |
| Predicted Positive | 88.9 % | 1.5 % | 0.16 % | 0.46 % |
| Precision | 0.07 % | 58.7 % | 95.2 % | 96 % |
| Recall | 100 % | 90.6% | 25 % | NA |

### 2) Sunkink

In order to detect a sunkink, model just needs to identify if the tracks are almost straight and parallel or not. This was first attempted using basic image processing –perspective transform, filtering and edge detection. This method worked quickly for detection of a sunkink in initial videos. In the processed image, shown in FIGURE 4: (A) ROI IN THE FRAME, (B) IMAGE AFTER PERSPECTIVE TRANSFORMED WITH GAUGE MEASUREMENTS FROM PROCESSED FRAME (C) PROCESSED ROI (D) AND (E) TRAINING IMAGES SIMULATED IN PAINT, (F) SUNKINK BEING DETECTED IN A VIDEOFIGURE 4 , thick white lines were identified as track. If the horizontal coordinates of left or right track varied quickly in a zig zag fashion, a sunkink was flagged. However, when this method was tried on multiple videos, it was observed, filtering techniques fail to detect track with the changes in ambience and lighting conditions. Even when filtering was made more robust, presence of switches, support rails, any straight bright object like pipes, fences, pillars or platform edge near track were creating distractions. Using fixes like accepting track coordinates only when track width is within limits or predicting track position based on where track was detected in the last frame helped to some extent but still the number of false positives in any new video were a lot.

It was felt after this experiment that to get a model which can be deployed in real world with new tracks in every run, conventional image processing might be the harder way to go, and simulating data to train a deep learning model might give better results. Positive scenarios of sunkinks were crudely simulated from good track images using MS Paint – surrounding ties were cropped and pasted on the track segment, then bent track was drawn on it by picking local color of track and its shadow. These images are shown in FIGURE 4. Basic dataset was made using just 50 positives and 100 negative images. 20% of these images were used as test dataset. For the training/testing, these images were further augmented using rotation, mirroring and image shifts. Trained models were cross validated on six 12-sec sunkink videos, professionally simulated using good videos obtained from a different terrain.

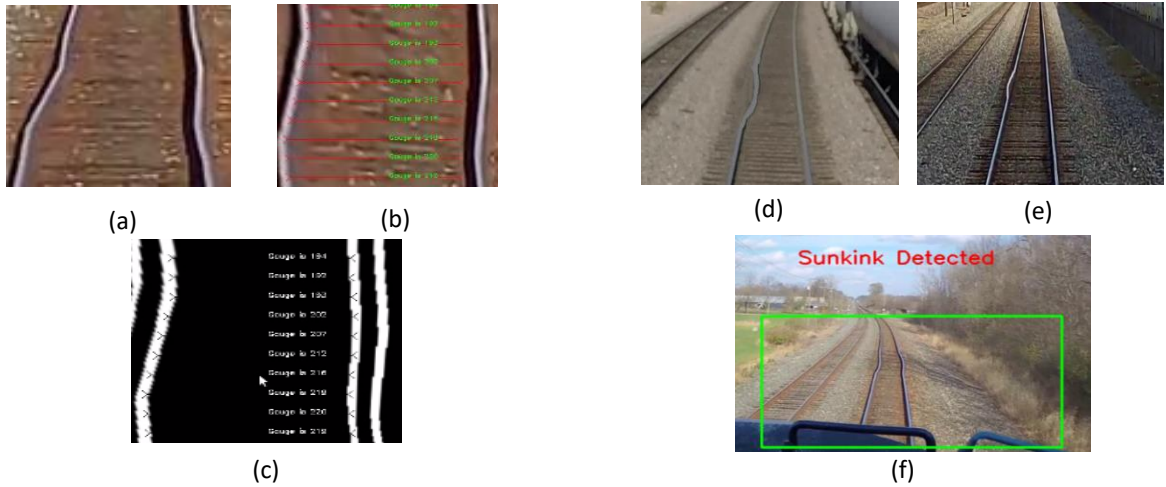

**Figure 4: (a) ROI in the frame, (b) Image after perspective transformed with Gauge measurements from processed frame (c) Processed ROI (d) and (e) Training images simulated in paint, (f) Sunkink being detected in a video**

Inception-v3 and Resnet-50 both the models, pretrained on ImageNet data were tried with a softmax layer at the end replacing the top layer of original model. Resnet model was also tried with top 10 network layers made trainable. Resnet model with extra trainable layers couldn't identify even 1 new sunkink resulting in an invalid precision, however resnet model with just top layer as trainable could detect 3 out of 6 sunkinks without giving any false positive. As very limited data was available, more trainable layers lead to an overfitted model. Inception model could detect all the sunkinks but also gave a false positive. It could also run real time on CPU. Inception architecture learns features with minimum number of trainable parameters, as a result of which it could work with even such a small dataset. These models were not getting distracted by switches or nearby platforms like conventional image processing methods. As sunkink is a severe rail defect which can cause derailment, running onboard is a critical requirement making inception a preferred choice among deep learning models.

**Table 2: Results for Sunkink Model**

| Parameters | Inception-v3 | ResNet 50 | ResNet 50 (More trainable layers) |
|---|---|---|---|
| Running Time (FPS) | 18 GPU, 6 CPU | 8 GPU, Not Running on CPU | 8 GPU, Not Running on CPU |
| Asset Level Accuracy | 6/6 | 3/6 | 0/6 |
| False +ves | 1 | 0 | 0 |
| Precision | 97.5 % | 100 % | NaN |

*B. Railway Assets*

As per Association of American Railroads regulations, companies need to maintain a GPS database of all their railroad assets which can be used by another operator driving a loco on their route. It helps in implementing dynamic speed limits in between two identified landmarks. Switch location known in advance helps the operator to stay alert in the region and onboard switch status detection at low speeds in yard can help prevent accidents. Signal color detection onboard can be used in operator assistance systems and hence avoiding having two operators onboard.

*1) Track Switches*

If seen in terms of image features to be recognized, switches are very similar to sunkinks. It's just that now a different shape of the track needs to be flagged. So, same architecture as sunkink model was tested at switches dataset. Simulating data for sunkinks is harder, but switches are available readily and labelling images for them is easier, so a much more robust model could be trained. 1408 positives and 4000 negatives were used. Data was further augmented with mirroring, image shifts and varied brightness. Similar to sunkinks, both the resnet 50 and inception v3 models, pretrained on imagenet data, with top layer modified, were tried for switches.

**Table 3. Results for Switch Detection**

| Parameters | ResNet 50 | Inception v3 | Inception v3 |
|---|---|---|---|
| Videos Parsed | 9 hrs | 9 hrs | 32 hrs |
| Precision | 53.8 % | 94.9 % | 95.6 % |
| Recall | 32.2 % | 83.2 % | NA |
| Performance Speed (fps) | CPU: 0.75 GPU: 8 | CPU: 10 GPU: 21 | CPU: 10 GPU: 21 |

Here also inception v3 model turns out to be preferable. The model could detect 83% of the switches in route with a precision value of 95%. It doesn't get confused with vipers on windshield and works even if it's raining. Recall couldn't be calculated as it needed labelling of 32 hrs of video to identify actual number of switches.

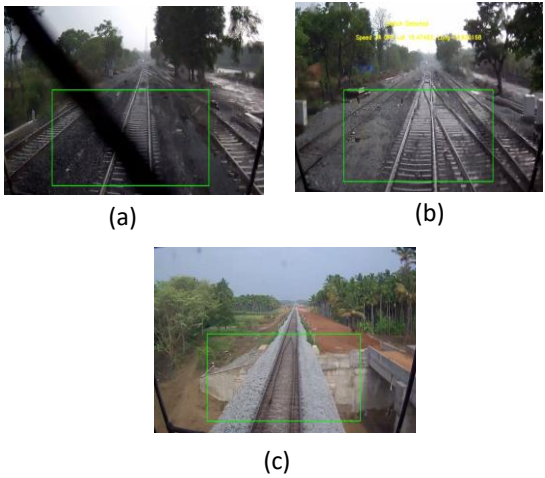

**Figure 5:** (a) and (b) shows switch detection working in rains, (c) shows model not getting confused with support rails

*2) Signal Color*

As there can be multiple signals present in a single frame, to be able to identify them individually so that their color can be detected, region based CNN models, pretrained on COCO dataset which provides bounding box labels as well has been used for signal detection. Another benefit was COCO dataset already includes road signals as one of the categories. Availability of all the labeled data set helped in model being able to adapt well to different regions with varied weather conditions.

Once the signal is detected, red and green color masks are applied on the image part inside the predicted bounding box to identify signal color. As the same signal is seen in multiple frames, even identification in any one of them is good. Thus, false positives were needed to be avoided but false negatives were fine. To handle this, if a signal was identified, but no proper color could be detected, it was ignored. This greatly helped in reducing false positives. Around 30 hrs of video spanning 700 kms of route was processed in multiple runs in sunny/cloudy weather conditions, afternoon/early morning timings.

**Table 4. Results for Signal Color Detection**

| Parameters | SSD | Faster R-CNN | Faster R-CNN |
|---|---|---|---|
| Videos Parsed | 1 hr | 1 hr | 30 hrs |
| Assets Detected | 11/51 | 47/51 | 411/440 |
| Asset Level Accuracy | 21.5 % | 92.1 % | 93.3 % |
| Precision | 98.4 % | 98.6 % | 99.4 % |
| Running Time (FPS) | CPU: 1, GPU: 15 | CPU: 0.1, GPU: 2 | CPU: 0.1, GPU: 2 |

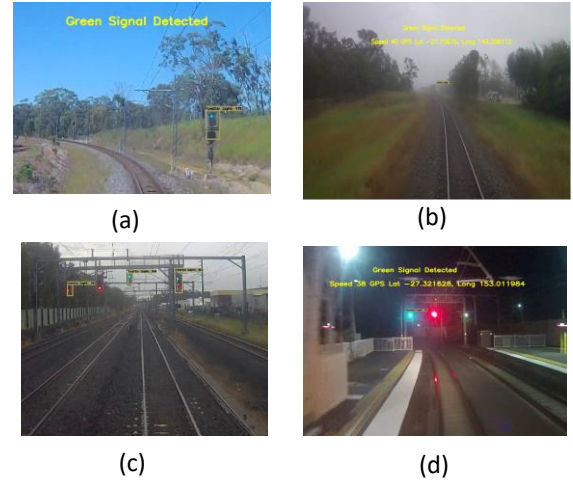

**Figure 6:** (a) Signal Detection in Metal box, (b) Overhead Signal Detection, (c) Signal Detection in Cloudy Weather, (d) Night time Signal Detection

Faster R-CNN model and Single Shot Detector (SSD) model Liu et al, 2016 were tried for this. SSD model could process. frames way faster than FRCNN but its detection accuracy was too low to be considered as a viable option. Using FRCNN, 30 hrs of videos were processed in which 93% of the signals present enroute could be identified with a precision of 99.4%. Detected signals included overhead signals, signals in metal cage, single light signals, triple light signals, etc. It was also observed, glare in sunny days was reducing accuracy and better performance was seen in early morning videos or cloudy weather. Though the model could detect daytime signals easily, for night time, asset level detection accuracy was as low as 20%. It was easier for it to detect night signals which were not too much refracted or blurred.

*C. Track Health Index*

To monitor the alerts for different track defects for a complete railway network in a region, a track health index has been proposed. All the track defects that can be identified will be given a weightage based on their severity. Defects like sunkink or broken track can cause derailment and hence will have higher weightage then something like loose ballast or vegetation overgrowth. THI value can be calculated as defined here:

$$THI = 1 - \text{MIN}(1, \Sigma_{\text{CLASSES}}(Confidence\ Percentage * Defect\ Weightage))$$

In the above equation, confidence percent denotes model's output probability for that defect to exist and defect weightage is the weight assigned to the particular defect based on its severity. For now, when we are working with just two defects, sunkink was assigned a weightage of 1 and loose ballast was assigned a weightage of 0.5. As more and more defects will be added in future, just the weights of different defects can be modified in the same formula. If for frame 1, a model is 80% sure that this is a case of loose ballast, THI for that frame would be 60%. For the category 2 defects, these THI values will be averaged over all the frames analyzed in a particular track segment to get the overall segment health. However, as category 1 defects do not spread over area and even a single occurrence is sufficient to cause derailment, the frames containing these defects will be flagged immediately.

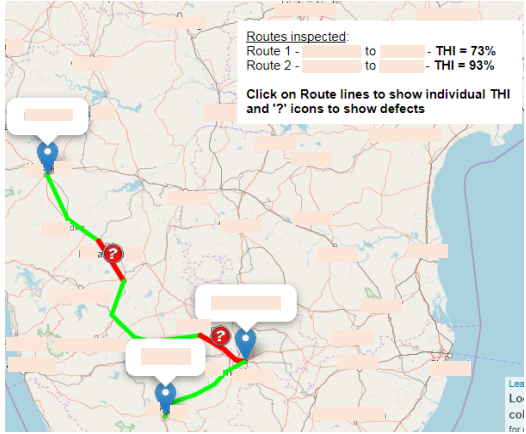

**Figure 7: Railway network with simulated Track Health Index mapped**

## V. CONCLUSION

From the experiments on rail domain data, it was observed that even in the cases where labeled data is not that readily available, using techniques like data augmentation and transfer learning can help in training deep learning models with decent accuracy. Models were trained to identify track defects like loose ballast and sunkinks and assets like switches and signals. From the performance of ResNet and Inception model in different cases, it was observed that Inception model could learn from a smaller dataset without overfitting. Multiple sized filters which are used in parallel in inception helps transmit more information to subsequent layers and hence tend to give better accuracy. However, if the feature to be detected is a little complex as was the case of loose ballast, deeper trainable model like ResNet work better. A comparison of SSD and Faster R-CNN models was also made for railway signals and it was observed that though SSD's lighter architecture makes it more desirable for real time detection but it's accuracy for new data is too low, making Faster R-CNN a more suited option despite higher computing resources requirement. Using all these detections, concept of track health index is proposed to enable monitoring and assess maintenance requirements for complete railway network.